%% file: sample-sigconf.tex
\documentclass[sigconf]{acmart}

\usepackage{multirow}
\usepackage{subfigure}
\usepackage{algorithm}
\usepackage[noend]{algorithmic}
\usepackage{xspace}
\usepackage{dsfont}
\usepackage{epsfig}
\usepackage{epstopdf}
\usepackage{booktabs}
\usepackage{indentfirst}
\usepackage{amsmath}
\usepackage{amsthm}
\usepackage{bm}
\usepackage{listings}
\usepackage{tabu}
\usepackage{longtable}
\usepackage{threeparttable}
\usepackage{marvosym}

\usepackage{tabularx}
\newtheorem{mydef}{Definition}

\newcommand{\team}{BUAA\_BIGSCity\xspace}

\newcommand{\ignore}[1]{}

\AtBeginDocument{%
  \providecommand\BibTeX{{%
    \normalfont B\kern-0.5em{\scshape i\kern-0.25em b}\kern-0.8em\TeX}}}

\setcopyright{acmcopyright}
\copyrightyear{2018}
\acmYear{2018}
\acmDOI{XXXXXXX.XXXXXXX}

\acmConference[Conference acronym 'XX]{Make sure to enter the correct
  conference title from your rights confirmation emai}{June 03--05,
  2018}{Woodstock, NY}
\acmPrice{15.00}
\acmISBN{978-1-4503-XXXX-X/18/06}

\begin{document}

\title{\team: Spatial-Temporal Graph Neural Network for Wind Power Forecasting in Baidu KDD CUP 2022}

\author{Jiawei Jiang}
\authornote{Both authors contributed equally to this research.}
\email{jwjiang@buaa.edu.cn}
\affiliation{%
  \institution{School of Computer Science and Engineering, Beihang University}
  \city{Beijing}
  \country{China}
}

\author{Chengkai Han}
\authornotemark[1]
\email{ckhan@buaa.edu.cn}
\affiliation{%
  \institution{School of Computer Science and Engineering, Beihang University}
  \city{Beijing}
  \country{China}
}

\author{Jingyuan Wang}
\authornote{*Corresponding author}
\email{jywang@buaa.edu.cn}
\affiliation{%
  \institution{School of Computer Science and Engineering, Beihang University}
  \city{Beijing}
  \country{China}
}

\renewcommand{\shortauthors}{Trovato and Tobin, et al.}

\begin{abstract}
In this technical report, we present our solution for the Baidu KDD Cup 2022 Spatial Dynamic Wind Power Forecasting Challenge. Wind power is a rapidly growing source of clean energy. Accurate wind power forecasting is essential for grid stability and the security of supply. Therefore, organizers provide a wind power dataset containing historical data from 134 wind turbines and launch the Baidu KDD Cup 2022 to examine the limitations of current methods for wind power forecasting. The average of RMSE (Root Mean Square Error) and MAE (Mean Absolute Error) is used as the evaluation score. We adopt two spatial-temporal graph neural network models, i.e., AGCRN and MTGNN, as our basic models. We train AGCRN by 5-fold cross-validation and additionally train MTGNN directly on the training and validation sets. Finally, we ensemble the two models based on the loss values of the validation set as our final submission. Using our method, our team \team achieves -45.36026 on the test set. We release our codes on Github~\footnote{\url{https://github.com/BUAABIGSCity/KDDCUP2022}} for reproduction.
\end{abstract}




\maketitle

\input{introduction.tex}

\input{methods.tex}

\input{experiments.tex}

\input{relatedwork.tex}


\input{conclusion.tex}


\bibliographystyle{ACM-Reference-Format}
\bibliography{sample-base}


\end{document}

%% file: introduction.tex
\section{Introduction}
Wind power is a rapidly growing source of clean energy. However, the dynamics and uncertainties of wind power can affect the grid network's system reliability. Therefore, accurate wind power forecasting is essential for reliable energy generation and smooth power system dispatching. 

Baidu KDD Cup 2022 Spatial Dynamic Wind Power Forecasting Challenge~\footnote{\url{https://aistudio.baidu.com/aistudio/competition/detail/152/0/introduction}} presents a unique Spatial Dynamic Wind Power Forecasting dataset: SDWPF, which contains the wind power data of 134 wind turbines from a wind farm over half a year with their relative positions and some dynamic context factors, such as wind speed, environment temperature, and turbine internal status. This Wind Power Forecasting (WPF) Challenge encourages the participants to develop effective models to accurately estimate the wind power supply of a wind farm at different time scales. The main difference between this challenge and past wind power prediction tasks is that the data provides the spatial distribution of turbines and various types of dynamic contextual information, so this competition is not a pure time series prediction task. The average of RMSE (Root Mean Square Error) and MAE (Mean Absolute Error) is used as the evaluation score.

In the literature, a series of works on wind power forecasting have emerged. The mainstream wind prediction methods can be divided into two main categories: classical statistical methods and machine learning (or deep learning) methods. The classical statistical methods are mainly some time series forecasting models, which are more dependent on the assumption of smoothness of the time series. Yunus et al. ~\cite{yunus2015arima} employ an autoregressive integrated moving average (ARIMA) based frequency-decomposed model to forecast short-term wind power. Recently, machine learning-based models have been widely used in wind power forecasting tasks. Wang et al. ~\cite{wang2019novel} use a support vector machine (SVM) to forecast short-term wind power. With the development of graph neural networks in recent years, the wind power forecasting task is no longer treated as a simple time series forecasting task. Because, in addition to the effects of wind speed, weather, and other factors, the output power of neighboring wind turbines interacts with each other and is spatially correlated. Graph neural networks (GNN) are suitable for capturing non-Euclidean graph structure relationships. For example, Li~\cite{li2022short} integrate the GNN and Deep Residual Network (DRN) for short-term wind power forecasting.

Our solution for this wind power forecasting task is also based on the graph neural network. Specifically, we extend and combine two spatial-temporal graph neural network models, i.e., AGCRN~\cite{bai2020adaptive} and MTGNN~\cite{wu2020connecting}, for wind power forecasting. The spatial-temporal graph neural network combines a graph neural network model and a temporal neural network model, such as the recurrent neural network (RNN) and 1D convolutional neural network (1D-CNN). In this way, the temporal neural network captures the temporal dependence between the wind power data of each single wind turbine, and the graph neural network captures the spatial correlation between the wind power data of multiple wind turbines. On top of these two base models, we introduce a relative geographic distance graph and a semantic distance graph, allowing the graph neural network to consider the interaction between wind turbines from different perspectives. We can achieve accurate wind power forecasting by capturing the dynamic spatial and temporal characteristics in wind power data. Our solution achieves an overall score of -45.36026.

%% file: methods.tex
\section{PRELIMINARIES}
In this section, we first introduce basic notations and preliminaries used in this paper. Then we formalize the problem of wind power forecasting.

\subsection{Notations and Definitions}

\begin{mydef}[Wind Turbine Farm]
A wind turbine farm contains $N$ wind turbines within a specific area, each constantly generating electricity. The Supervisory Control and Data Acquisition (SCADA) system of the wind farm continuously records the power generated by each wind turbine in the wind farm at a fixed time interval.
\end{mydef}

\begin{mydef}[Wind Power Tensor]
In this paper, we use $\bm{X}_t \in \mathbb{R}^{N \times C}$ to denote the observation at time $t$ of $N$ wind turbines in the wind farm, where $C$ is the length of the observation vector. The observation includes active wind power, wind speed, etc., as described in Section \ref{data}. In addition, we use $\bm{X} = (\bm{X}_1, \bm{X}_2, \cdots, \bm{X}_T) \in \mathbb{R}^{T \times N \times C}$ to denote the wind power tensor of all wind turbines at $T$ time slices.
\end{mydef}

\subsection{Problem Formalization}
The wind power forecasting task aims to predict the wind power supply of a wind farm in the future time given the historical observations. Formally, given the tensor $\bm{X}$ observed on a wind farm, our goal is to learn a mapping function $f$ from the observations of the previous $T$ steps to predict the wind power supply of the future $T'$ steps as:
\begin{equation}\label{eq:problem_def}
[\bm{X}_{(t-T+1)}, \cdots, \bm{X}_t] \stackrel{f}{\longrightarrow} [\hat{\bm{X}}_{(t+1)}, \cdots, \hat{\bm{X}}_{(t+T')}].
\end{equation}
Note that the prediction result tensor $\hat{\bm{X}} \in \mathbb{R}^{T' \times N \times 1}$ contains only one-dimensional features, i.e., the active power generated by each wind turbine.

\section{Methods}
In this section, we describe the two models used in this challenge.


\begin{figure}[t]
    \centering
    \includegraphics[width=0.9\columnwidth]{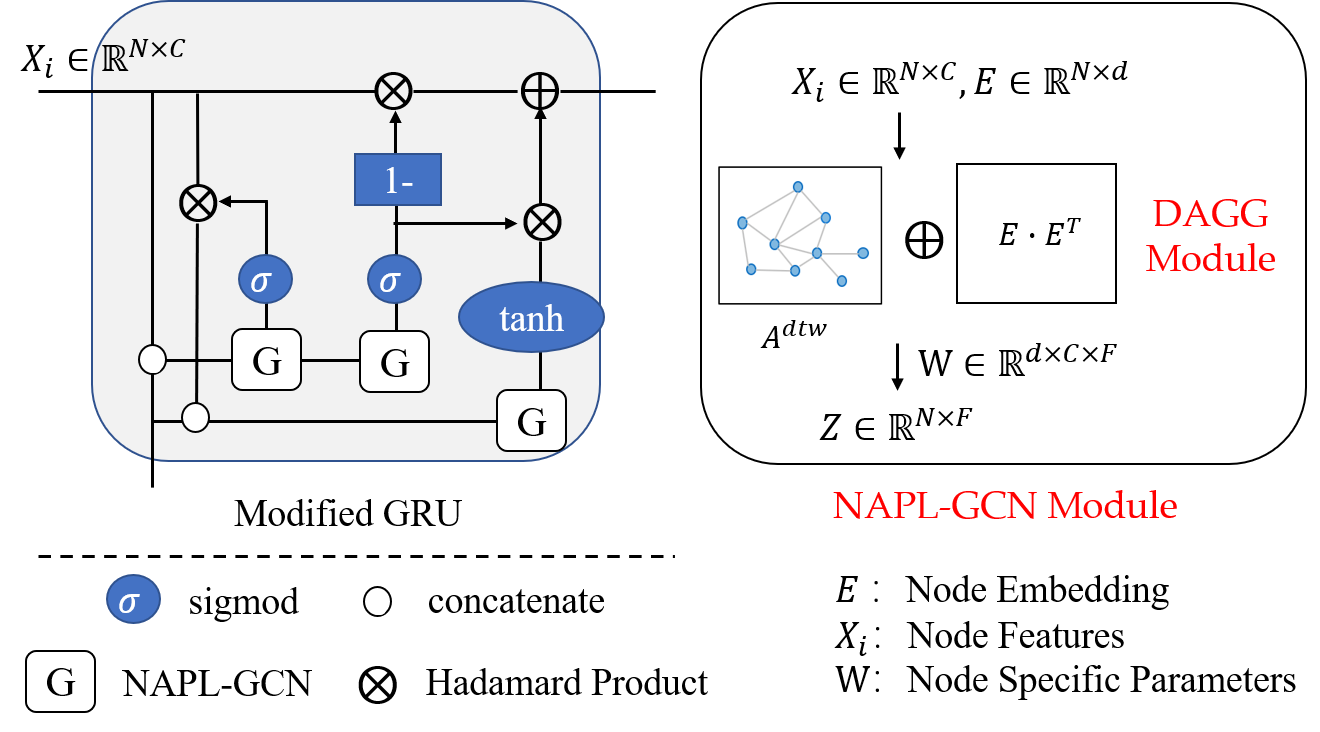}
    \caption{Overall Structure of AGCRN}
    \label{fig:agcrn}
\end{figure}

\subsection{AGCRN}

Adaptive Graph Convolutional Recurrent Network (AGCRN) is proposed in ~\cite{bai2020adaptive} NIPS2020. AGCRN consists of two adaptive modules to enhance Graph Convolution Network (GCN) and proposes the NAPL-GCN module: (1) a Node Adaptive Parameter Learning (NAPL) module to capture node-specific patterns; (2) a Data Adaptive Graph Generation (DAGG) module to infer the inter-dependencies among different time series automatically. To capture temporal correlations in the data, AGCRN replaces the MLP layers of the Gated Recurrent Unit (GRU)~\cite{cho2014learning} with the NAPL-GCN module to capture both node-specific spatial and temporal correlations in the time series. The overall structure of AGRCN is shown in Figure~\ref{fig:agcrn}. Here we modify the DAGG module and add a pre-defined semantic distance graph of all turbines to directly capture the semantic relationships between the time series of different wind turbines. We stack several AGCRN layers as an encoder to capture the spatial-temporal correlations in the wind power data. Finally, we obtain the predicted wind power supply time series of all wind turbines by applying a single linear transformation to project the representation generated by the encoders.

\subsubsection{Node Adaptive Parameter Learning (NAPL)}
GCN is widely used to capture spatial correlations in data. According to ~\cite{kipf2016semi}, graph convolution operation can be well approximated as:
\begin{equation}
\bm{Z}=(\bm{I}_N + {D}^{-\frac{1}{2}}{A}{D}^{-\frac{1}{2}})\bm{X}_{i}\bm{\Theta}+\bm{b}
\end{equation}
where ${A}\in\mathbb{R}^{N\times N}$ is the adjacency matrix of the graph, $\bm{D}$ is the degree matrix, $\bm{X}_{i}\in\mathbb{R}^{N\times C}$ and $\bm{Z}\in\mathbb{R}^{N\times F}$ are input and output of the GCN layer, $\bm{\Theta}\in\mathbb{R}^{C\times F}$ and $\bm{b}\in\mathbb{R}^{F}$ are the learnable weights and bias respectively. However, it is difficult to model diverse patterns of all turbines with the shared $\bm{\Theta}$ and $\bm{b}$. To learn \textit{node-specific} patterns, the NAPL module maintains a unique parameter space for each turbine. Besides, instead of directly learning $\bm{\Theta}\in\mathbb{R}^{N\times C\times F}$, which is too large to optimize, this module learns two smaller parameter matrix: a node embedding matrix $\bm{E}\in\mathbb{R}^{N\times d}$ and a weight pool $\tilde{\bm{W}}\in\mathbb{R}^{d\times C\times F}$, where $d$ is the embedding dimension and $d\ll N$. In this way, $\bm{\Theta} = \bm{E}\tilde{\bm{W}}$. We use the same operation for $\bm{b}\in\mathbb{R}^{N\times F}$, i.e., $\bm{b} = \bm{E}\tilde{\bm{b}}, \tilde{\bm{b}} \in \mathbb{R}^{d\times F}$. Finally, the NAPL-GCN module is calculated as:
\begin{equation}
\bm{Z}=(\bm{I}_N + {D}^{-\frac{1}{2}}{A}{D}^{-\frac{1}{2}})\bm{X}_i\bm{E}\tilde{\bm{W}}+\bm{E}\tilde{\bm{b}}.
\end{equation}

\subsubsection{Data Adaptive Graph Generation (DAGG)}
The DAGG module automatically infers the spatial dependencies between each pair of wind turbines for the graph convolution operation. First, this module uses the same learnable node embedding $\bm{E}\in\mathbb{R}^{N\times d}$ as the NAPL-GCN module for all wind turbines, where each row of $\bm{E}$ represents the node embedding of a wind turbine. We directly generate ${D}^{-\frac{1}{2}}{A}{D}^{-\frac{1}{2}}$ by multiplying $\bm{E}$ and $\bm{E}^T$ as:
\begin{equation}
 {D}^{-\frac{1}{2}}{A}{D}^{-\frac{1}{2}}={\rm softmax}({\rm ReLU}(\bm{E}\cdot\bm{E}^T)),
\end{equation}
where ${\rm softmax}$ and ${\rm ReLU}$ are activation functions. During training, the node embeddings $\bm{E}$ will be updated automatically to learn the hidden spatial dependencies among all wind turbines. Then, the DAGG enhanced NAPL-GCN can be expressed as:
\begin{equation}
\bm{Z}=(\bm{I}_N + {\rm softmax}({\rm ReLU}(\bm{E}\cdot\bm{E}^T)))\bm{X}_i\bm{\Theta}+\bm{b}.
\end{equation}


\begin{figure}[t]
    \centering
    \subfigure[Semantic Distance Graph, $M$=5]{
        \includegraphics[width=0.45\columnwidth, page=1]{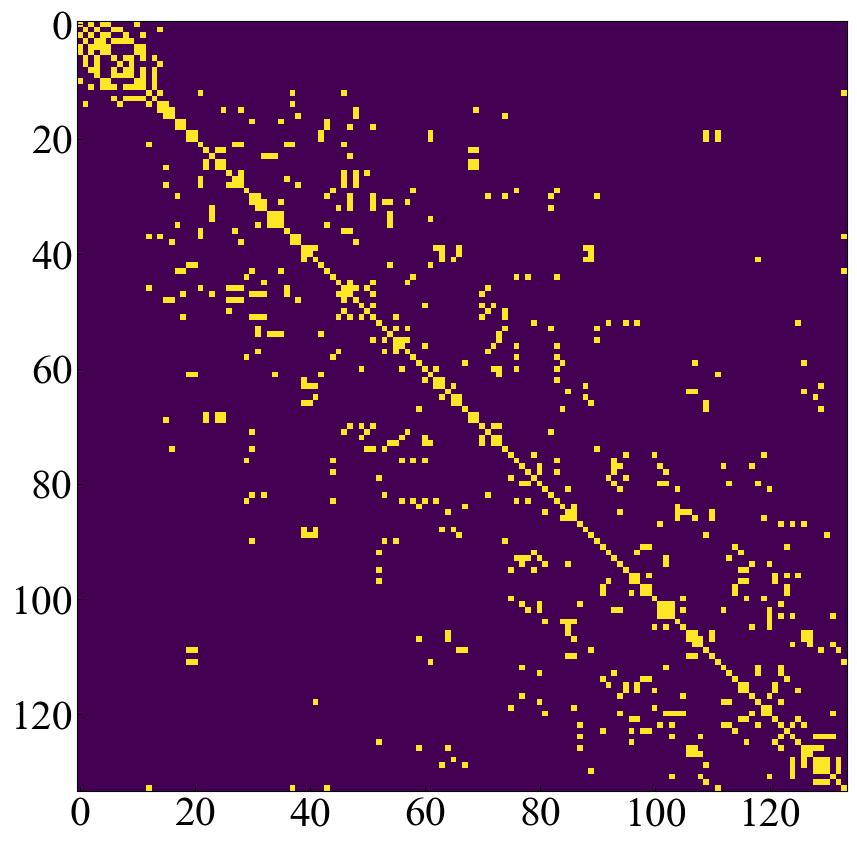}
        \label{fig:dtw}
    }
    \subfigure[Geographic Distance Graph, $\epsilon=0.8$]{
        \includegraphics[width=0.45\columnwidth, page=2, page=1]{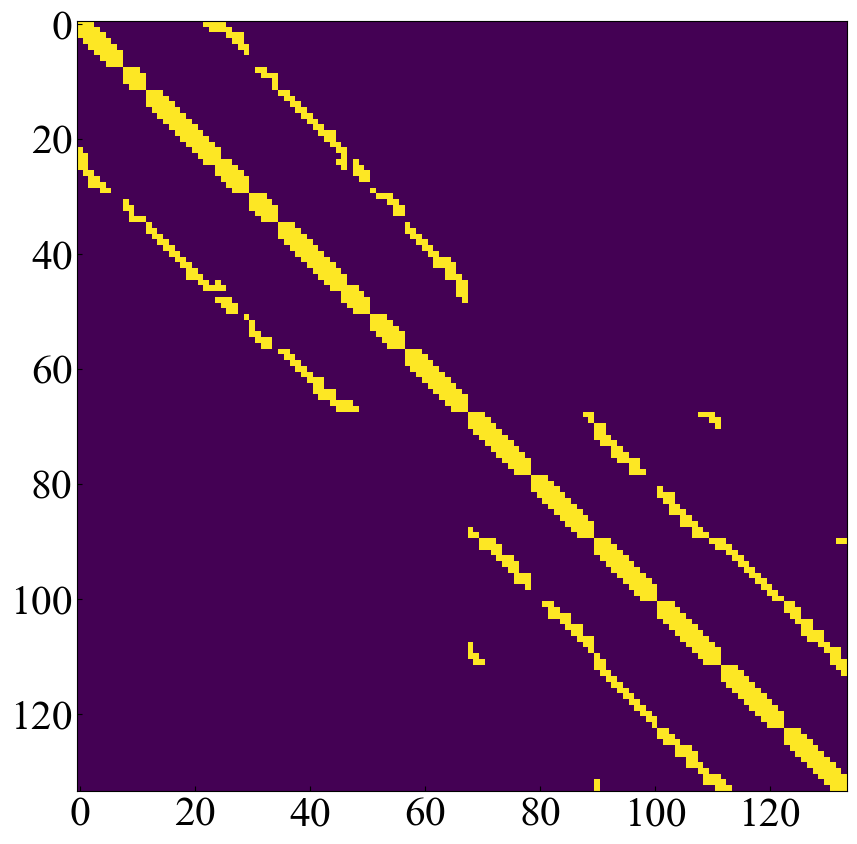}
        \label{fig:geo}
    }
    \caption{Different Adjacency Graphs}
    \label{fig:graph}
\end{figure}

\subsubsection{Semantic Distance Graph}
The DAGG enhanced NAPL-GCN module can adaptively learn the spatial dependencies among all wind turbines and the \textit{node-specific} patterns among them. As a multivariate time series forecasting task, there is not only geographical distance correlation between multiple time series but also semantic similarity, i.e., time series similarity. Therefore, a semantic distance graph is also considered for the graph convolution operation. We compute the Dynamic Time Warping (DTW) distance ~\cite{yi1998efficient} between the historical wind power supply time series of each pair of wind turbines and obtain a similarity matrix $D^{dtw}$. Then, for each wind turbine, $M$ wind turbines with the top-$M$ smallest DTW distance are selected as neighbors to obtain a binary semantic distance graph $A^{dtw}$, shown in Figure~\ref{fig:dtw}. Finally, we append $A^{dtw}$ to the DAGG enhanced NAPL-GCN module as follows:
\begin{equation}\label{agcrn}
\bm{Z}=(\bm{I}_N + {\rm softmax}({\rm ReLU}(\bm{E}\cdot\bm{E}^T)) + {D}^{-\frac{1}{2}}{A^{dtw}}{D}^{-\frac{1}{2}})\bm{X}_i\bm{\Theta}+\bm{b},
\end{equation}
where ${D}$ is the degree matrix of ${A^{dtw}}$, i.e., $D_{i, i} = \sum_j A^{dtw}_{i,j}$. Equation (\ref{agcrn}) is the complete formulation of the graph convolution module of the modified AGCRN model to capture spatial correlation. To further capture temporal correlation, the AGCRN model replaces the MLP layers in the GRU model with the graph convolution module described above. We stack multiple AGCRN layers and obtain the prediction results of the model by a linear transformation on top of this.

\begin{figure}[t]
    \centering
    \includegraphics[width=0.9\columnwidth]{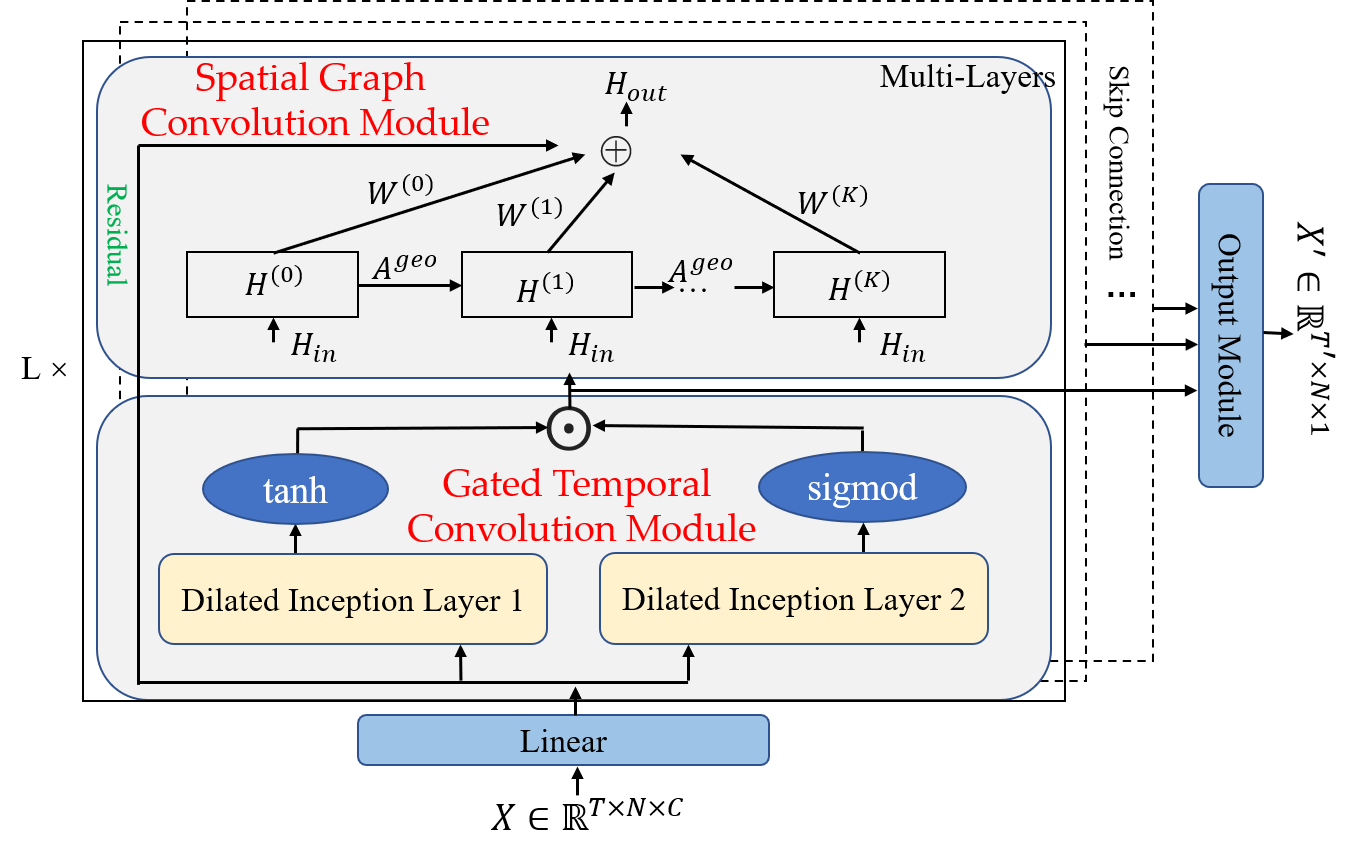}
    \caption{Overall Structure of MTGNN}
    \label{fig:mtgnn}
\end{figure}

\subsection{MTGNN}
Multivariate Time Series Forecasting Graph Neural Networks (MTGNN) is proposed in ~\cite{wu2020connecting} KDD2021. As shown in Figure~\ref{fig:mtgnn}, the overall structure of MTGNN consists of multiple sequentially connected gated temporal convolutional modules and spatial graph convolutional modules, with additional residual connections to make the model easy to train. Skip connections are added after each temporal convolution module to get the output hidden features. Finally, the output module consists of two 1*1 convolutions is used to project the hidden features to the desired output dimension, i.e., the length of the predicted time window $T'$. Here we modify the adaptive graph structure learning module in MTGNN and instead use a graph based on the relative geographic distance between wind turbines.


\subsubsection{Geographic Distance Graph}
The relative geographic relationship between wind turbines, such as the upstream vs. downstream relationship, has a significant impact on the power generated by wind turbines. Therefore, we calculate the Euclidean distance between two turbines based on the coordinates of each turbine given in the data and obtain the geographic distance matrix $D^{geo}$. We believe that the closer the turbines are to each other, the greater the influence. Therefore, we define the following weighting formula to obtain the geographic distance graph $A^{geo}$, which is shown in Figure~\ref{fig:geo}:
\begin{equation}
A^{geo}_{i,j} =
\left\{  
\begin{aligned}  
& 1, & \exp(-\frac{{(D^{geo}_{i,j})}^2}{\sigma^2}) \geq \epsilon \\  
& 0, & \exp(-\frac{{(D^{geo}_{i,j})}^2}{\sigma^2}) < \epsilon,  
\end{aligned}  
\right. 
\end{equation}
where $D^{geo}_{i,j}$ represents the distance between turbine $i$ and turbine $j$, the $\sigma$ is the standard deviation of distances, and $\epsilon$ is the threshold to control the sparsity of the geographic distance graph.

\subsubsection{Spatial Graph Convolution Module}
The graph convolution module used here is a mixhop propagation layer. Given the geographic distance graph ${A^{geo}}$, the mix-hop propagation layer consists of two steps: (1) the information propagation step and (2) the information selection step. There are $K$ steps of information propagation, each of which is calculated as follows:
\begin{equation}
\bm{H}^{(k)} = \beta \bm{H}_{in} + (1 - \beta) \tilde{{A}} \bm{H}^{(k-1)},
\end{equation}
where $\bm{H}_{in}$ is the input hidden states of this layer, i.e. the output of the previous layer, $\bm{H}^{(k)}$ is the output after the $k$-th propagation, $\beta$ is the hyperparameter which controls the proportion of input features retained, and $\tilde{A}$ is the Laplace matrix of the geographic distance graph, $\tilde{A} = \tilde{D}^{-1}(A^{geo}+I), \tilde{D}_{i,i} = 1 + \sum_{j}A^{geo}_{i,j}$.

The information selection step is a weighted aggregation of the results of the $K$-step propagation so that the model can make an adaptive selection. The information selection step is defined as follows:
\begin{equation}
    \bm{H}_{out} = \sum_{i=0}^K \bm{H}^{(k)} \bm{W}^{(k)},
\end{equation}
where $K$ is the total steps of information propagation, $\bm{W}^{(k)}$ are learnable parameters, $\bm{H}_{out}$ is the output hidden states of the mixhop propagation graph convolution module. In general, this mixhop propagation graph convolution, which is a balance between the nodes' local information and the neighborhood information, avoids the over-smoothing problem of the graph convolution model~\cite{chen2020measuring}.

\subsubsection{Gated Temporal Convolution Module}
The temporal convolution module consists of two dilated inception layers. The first dilated inception layer is followed by a tangent hyperbolic (tanh) activation function as a filter, and the other is followed by a sigmoid activation function as a gate. The process of calculation is as follows:
\begin{equation}
    \bm{T}_{out} = {\rm tanh}(\bm{T}_1) \odot {\rm sigmod}(\bm{T}_2)
\end{equation}
where $\odot$ indicates the Hadamard product and $\bm{T}_1, \bm{T}_2$ are the output of the two dilated inception layers, respectively.

The dilated inception layer is implemented with 1D convolutional filters to deal with the time series data. Besides, to increase the receptive field to handle longer time series and reduce model complexity, we use a dilated convolution whose receptive field grows exponentially by two as the number of hidden layers increases. Besides, to discover temporal patterns with various ranges, we use four different sizes of filters, including $1 \times 2$, $1 \times 3$, $1 \times 6$, and $1 \times 7$. The outputs of the four filters are truncated to the same length according to the largest filter and concatenated across the channel dimension to get the model outputs.

%% file: experiments.tex
\section{Experiments}

\subsection{Datasets} \label{data}

The SDWPF dataset~\footnote{\url{https://aistudio.baidu.com/aistudio/competition/detail/152/0/datasets}} is collected from Supervisory Control and Data Acquisition (SCADA) systems on wind turbines of a wind farm owned by Longyuan Power Group~\cite{zhou2022sdwpf}. The SCADA data are sampled every 10 minutes from each wind turbine in the wind farm, which consists of 134 wind turbines and the total time span of the data is 245 days. The detailed information of the SDWPF dataset is shown in Table~\ref{tab:statistics}.

\begin{table}
  \caption{Statistics of the SDWPF dataset}
  \label{tab:statistics}
  \begin{tabular}{ccccl}
    \toprule
    Days & Interval & \# of columns & \# of turbines & \# of records\\
    \midrule
    245 & 10 minutes & 13 & 134 & 4,727,520\\
  \bottomrule
\end{tabular}
\end{table}

This dataset contains important external features, such as wind speed and external temperature, which can influence wind power generation, as well as critical internal features, such as the inside temperature, that indicate the operating status of each wind turbine. A detailed introduction of the main attributes of the SDWPF dataset is listed in Table~\ref{tab:column}.

\begin{table*}
  \caption{Column names and their specifications of the SDWPF dataset}
  \label{tab:column}
  \begin{tabular}{ccl}
    \toprule
    Column & Column Name & Specification\\
    \midrule
    1 & TurbID & Wind turbine ID \\
    2 & Day & Day of the record \\
    3 & Tmstamp & Created time of the record\\
    4 & Wspd (m/s) & The wind speed recorded by the anemometer \\
    5 & Wdir ($^{\circ}$) & The angle between the wind direction and the position of turbine nacelle \\
    6 & Etmp ($^{\circ}$C) & Temperature of the surounding environment \\
    7 & Itmp ($^{\circ}$C) & Temperature inside the turbine nacelle \\
    8 & Ndir ($^{\circ}$) & Nacelle direction, i.e., the yaw angle of the nacelle \\
    9 & Pab1 ($^{\circ}$) & Pitch angle of blade 1 \\
    10 & Pab2 ($^{\circ}$) & Pitch angle of blade 2 \\
    11 & Pab3 ($^{\circ}$) & Pitch angle of blade 3 \\
    12 & Prtv (kW) & Reactive power \\
    13 & Patv (kW) & Active power (target variable) \\
    \bottomrule
  \end{tabular}
\end{table*}

\begin{figure}[t]
    \centering
    \includegraphics[width=0.5\columnwidth]{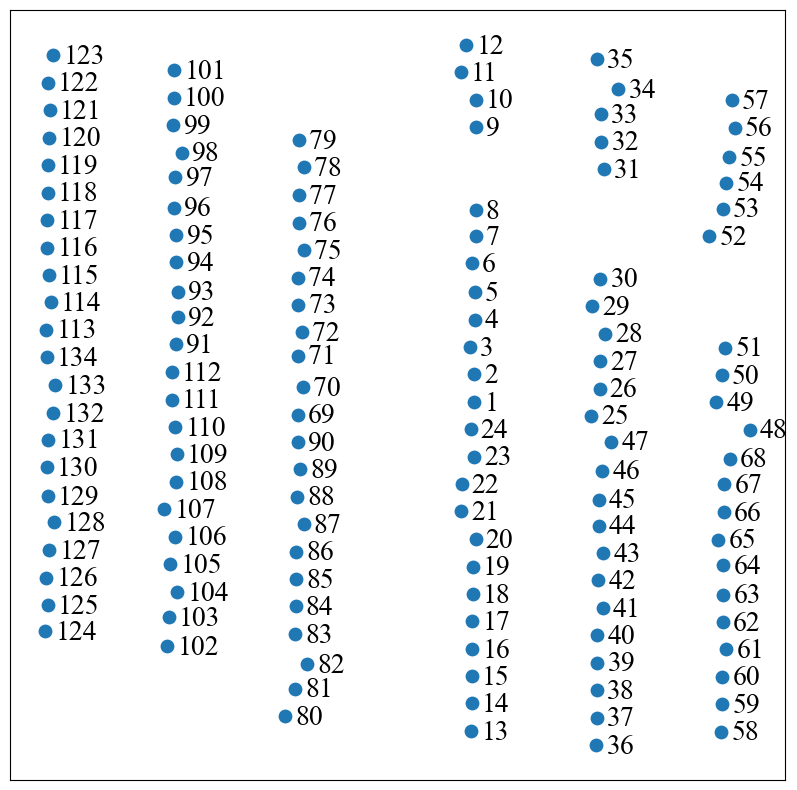}
    \caption{Spatial Distribution of Wind Turbines}
    \label{fig:spatial}
\end{figure}

In addition, the SDWPF dataset provides the relative position for all wind turbines. Figure~\ref{fig:spatial} characterizes the spatial distribution of the 134 wind turbines in the wind farm.

\subsection{Experimental Settings}

\subsubsection{Dataset Processing}
To make full use of the SDWPF dataset, we apply some feature engineering approaches to the dataset. First, we select useful features for the model training and remove irrelevant or redundant features according to correlation coefficient scores~\cite{benesty2009pearson}. For attributes listed in Table~\ref{tab:column} we select the following five features, including ${Wspd}$, ${Etmp}$, ${Itep}$, ${Prtv}$ and ${Patv}$. Second, we construct a new feature $\Delta Patv$ based on $Patv$. At time step $t_0$, $\Delta Patv$ is the difference between $Patv$ at time step $t_0$ and $Patv$ at time step $t_0 - 1$. In addition, before training, we use Z-score normalization on the dataset to standardize the data inputs. Finally, following the official release of the baseline code~\footnote{\url{https://github.com/PaddlePaddle/PGL/tree/main/examples/kddcup2022/wpf_baseline}}, we add a data argument strategy. During the model training, we fuse the historical time series of each wind turbine with data from different periods of the same wind turbine to enrich the data samples and mitigate the occurrence of overfitting.

\subsubsection{Evaluation Metrics} \label{metric}

Wind power forecasting aims to predict the wind farm's time series of wind power. However, due to the underlying outliers in the SDWPF dataset, we evaluate the prediction results for each wind turbine and then sum the prediction scores as the final score of the model. In this wind power forecasting challenge, our target is to predict a future length-288 wind power supply time series, and the average of RMSE (Root Mean Square Error) and MAE (Mean Absolute Error) is used as the evaluation score. Therefore, at the time step $t_0$, the evaluation score $s_{t_0}^i$ for wind turbine $i$ is defined as:
\begin{equation}
    s_{t_0}^i=\frac{1}{2}\left(\sqrt{\frac{\sum_{j=1}^{288}(\bm{X}_{t_0+j}^i-\hat{\bm{X}}_{t_0+j}^i)^2}{288}}+\frac{\sum_{j=1}^{288}|\bm{X}_{t_0+j}^i-\hat{\bm{X}}_{t_0+j}^i|}{288}\right),
\end{equation}
where $\bm{X}_{t_0+j}^i$ is the ground-truth active power ($Patv$) of wind turbine $i$ and $\hat{\bm{X}}_{t_0+j}^i$ is the predicted active power of wind turbine $i$ at time step $t_0+j$. At time step $t_0$, the final score $S_{t_0}$ is the sum of the evaluation score on all 134 wind turbines.

As for the outliers, we introduce a few specific treatments when using this data. This section is the same as in the official introduction technical report~\cite{zhou2022sdwpf}.

\textbf{Zero values.} There are some reactive power and active power which are smaller than zeros. We simply treat all the values which are smaller than 0 as 0.

\textbf{Missing values.} Some values at some time are not collected from the SCADA system and will not be used for evaluating the model.

\textbf{Unknown values.} At some times, wind turbines stop generating electricity for external reasons, such as wind turbine modification and/or active scheduling of power supply to avoid grid overload. In these cases, the actual generated active power of the wind turbine is unknown. These unknown values will not be used to evaluate the model. In this challenge, two conditions are introduced to determine whether the target variable is unknown:
\begin{itemize}
\item If at time step $t_0$, there are $Patv \leq 0$ and $W_{spd} > 2.5$, then the actual active power $Patv$ of this wind turbine at time step $t_0$ is unknown;
\item If at time step $t_0$, there are $Pab1 > 89^{\circ}$ or $Pab2 > 89^{\circ}$ or $Pab3 > 89^{\circ}$, then the actual active power $Patv$ of this wind turbine at time step $t_0$ is unknown.
\end{itemize}

\textbf{Abnormal values.} If there are any abnormal values in any column of the data record, these values will not be used to evaluate the model. In this challenge, we define two rules to identify abnormal values:
\begin{itemize}
\item If at time step $t_0$, $Ndir > 720^{\circ}$ or $Ndir < -720^{\circ}$, then the actual active power $Patv$ of this wind turbine at time step $t_0$ is abnormal;
\item If at time step $t_0$, $Wdir > 180^{\circ}$ or $Wdir < -1808{\circ}$, then the actual active power $Patv$ of this wind turbine at time step $t_0$ is abnormal.
\end{itemize}

\subsubsection{Training Loss}
We choose to use the Huber loss~\cite{huber1992robust} as the training loss function of the two models because the Huber loss is less sensitive to outliers than the squared error loss, which can be expressed as follows:
\begin{equation}
\mathcal{L}(\hat{\bm{X}}, \bm{X}) = 
\left\{
    \begin{aligned}
    & \frac{1}{2}(\hat{\bm{X}} - \bm{X}) ^ 2 & |\hat{\bm{X}} - \bm{X}| \leq \delta \\
    & \delta|\hat{\bm{X}} - \bm{X}| - \frac{1}{2}\delta^2 & |\hat{\bm{X}} - \bm{X}| > \delta \\
    \end{aligned}\right.,
\end{equation}
where $\delta$ is the threshold parameter to control the sensitivity of squared error, $\hat{\bm{X}}$ represents the predicted active power of all wind turbines, and $\bm{X}$ represents the ground-truth active power of all wind turbines. Note that all the predicted and ground-truth values have been treated with rules for outliers in Section \ref{metric}.

\subsubsection{Training and Model Settings}

All experiments are conducted on Ubuntu 18.04 with an NVIDIA GeForce 3090 GPU. We predict a future length-288 wind power supply time series with the past length-144 wind power supply time series, i.e., $T=144, T'=288$. We implement two models based on the PyTorch~\footnote{\url{https://pytorch.org/}} framework. We train our model using an Adam~\cite{kingma2014adam} optimizer with a learning rate of 0.001. The batch size is 32, and the training epoch is 30. A clipping gradient mechanism is used to stabilize the gradient for training better, and the clipping gradient is set to 5. As for the Huber loss, the threshold parameter $\delta$ is 5.

For training AGCRN, we stack two AGCRN layers and set the hidden unit to 64 for all the AGCRN cells. The embedding dimension $d$ of each node is 10. In the semantic distance graph $A_{dtw}$, the number of similar wind turbines $M$ is 5. For training MTGNN, we stack three sequentially connected gated temporal convolutional modules and spatial graph convolutional modules. The hidden dimension is 32, and the skip connection dimension is 64. The number of information propagation steps $K$ is 2. The proportion of input features retained in the Spatial Graph Convolution Module $\beta$ is 0.05, and the dilation exponential factor in the Gated Temporal Convolutional Module is 2. In the geographic distance graph $A_{geo}$, the threshold to control the sparsity $\epsilon$ is 0.8.

\subsection{Empirical Results}

Table~\ref{tab:performance} summarizes the performance of models used in the final submission. A 5-fold cross-validation strategy is adopted for training AGCRN. As shown in Table~\ref{tab:performance}, we divide all the data into five equal folds, then AGCRN is trained using the other four folds and validated on the selected fold. MTGNN is directly trained on the training set containing data of 214 days and validated on the validation set containing data of 31 days.

\begin{table}
  \caption{Model Performance}
  \label{tab:performance}
  \begin{tabular}{c|c|l}
    \toprule
    Model & Fold & Val Loss\\
    \midrule
    \multirow{5}*{AGCRN} & 0 & 0.3218\\
    ~ & 1 & 0.4953\\
    ~ & 2 & 0.3570\\
    ~ & 3 & 0.2988\\
    ~ & 4 & 0.4085\\
    \hline
    MTGNN & - & 0.2512\\
  \bottomrule
\end{tabular}
\end{table}

After training, we leverage a naive weighted average to ensemble the output of 5 AGCRN models according to the performance of valid loss. Specially, we perform a weighted fusion of the prediction results of the 5 AGCRN models based on the reciprocals of valid losses, i.e., [$0.3218^{-1}$, $0.4953^{-1}$, $0.3570^{-1}$, $0.2988^{-1}$, $0.4085^{-1}$]. In this way, the larger the validation set loss, the smaller the proportion of models in the ensemble model. After obtaining the ensembled AGCRN model, we integrate the ensembled AGCRN model and MTGNN model again according to the ratio of 4:6 and obtain the final model prediction results. Using this method, we achieve -45.36026 on the test set finally.

%% file: relatedwork.tex
\section{Related Work}


The mainstream wind power forecasting methods can be divided into two main categories: classical statistical methods and machine learning (or deep learning) methods. The classical statistical methods are mainly some time series forecasting models. For example, Han et al.~\cite{han2017non} utilized the autoregressive moving average (ARMA) model to forecast wind power. Recently, machine learning-based models have been widely used in wind power forecasting tasks. Wang et al. ~\cite{wang2019novel} used a support vector machine (SVM) to forecast short-term wind power. 

Deep learning-based algorithms are currently the dominant algorithms for wind power forecasting. Chen et al. ~\cite{chen2021deep} used a joint model composed of Long Short-Term Memory Network (LSTM)\cite{hochreiter1997long} and Convolutional Neural Network (CNN) to forecast wind power for multi-turbines. Liu et al. ~\cite{liu2018smart} also integrated a Long Short-Term Memory Network (LSTM) with variational mode decomposition for short-term wind power forecasting. In recent years, with the development of graph neural networks (GNN), wind power forecasting is no longer treated as a simple time series forecasting task. Researchers have begun using graph neural networks to capture spatial correlations between wind turbines. For example, Khodayar et al. ~\cite{khodayar2018spatio} used LSTM to extract temporal features and a first-order approximation of spectral graph convolution to capture spatial features for wind power forecasting. Li~\cite{li2022short} integrated the GNN and Deep Residual Network (DRN) for short-term wind power forecasting. Bentsen et al. ~\cite{bentsen2022wind} proposed a modular framework using attention-based graph neural networks for wind power forecasting.

%% file: conclusion.tex
\section{Conclusion}

In this technical report, we present our solution for the Baidu KDD Cup 2022 Spatial Dynamic Wind Power Forecasting Challenge. This competition provides a large-scale dataset of 134 wind turbine historical data, with the relative position of the turbines and various dynamic contextual information. We extend two spatial-temporal graph neural network models, i.e., AGCRN and MTGNN, as our basic models. These models capture the data's dynamic temporal and spatial correlations, enabling accurate wind power forecasting. We train AGCRN by 5-fold cross-validation and additionally train MTGNN directly on the training and validation sets. Finally, we ensemble the two models as our final submission. Using our method, our team \team achieves -45.36026 on the test set.